\documentclass[english]{ccdconf}
\usepackage{amsmath}
 \usepackage{amssymb}
\usepackage[scaled=1.0]{helvet}
\usepackage{times}
\usepackage{graphicx}
\usepackage{subfigure}
\usepackage{parskip}
\usepackage{multirow}
\usepackage{enumitem} 
\usepackage[lined,boxed,commentsnumbered, ruled]{algorithm2e}

\begin{document}

\title{Object Servoing of Differential-Drive Robots}
\author{Weibin Jia\aref{wrjs}, Wenjie Zhao\aref{wrjs},
      Zhihuan Song\aref{gkll} and Zhengguo Li\aref{lsxx}}

\affiliation[wrjs]{School of Aeronautics and Astronautics, Zhejiang University, Zhejiang  310027, China
        \email{j@zju.edu.cn, zhaowenjie8@zju.edu.cn}}
\affiliation[gkll]{School of Control Science and Engineering, Zhejiang University, Zhejiang  310027, China
        \email{songzhihuan@zju.edu.cn}}
\affiliation[lsxx]{SRO Department of  Institute for Infocomm Research, 1 Fusionopolis Way, 138632, Singapore
        \email{ezgli@i2r.a-star.edu.sg}}

\maketitle

\begin{abstract}
Due to possibly changing pose of a movable object and nonholonomic constraint of a differential-drive robot, it is challenging to design an object servoing scheme for the differential-drive robot to asymptotically park at a predefined relative pose to  the movable object. In this paper, a novel object servoing scheme is designed for the differential-drive robots. Each on-line relative pose is first estimated by using feature points of the moveable object and it serves as the input of an object servoing friendly parking controller. The linear velocity and angular velocity are then determined by the parking controller. Experimental results validate the performance of the proposed object servoing scheme. Due to its low on-line computational cost, the proposed scheme can be applied for last mile delivery of differential-drive robots to movable objects.
\end{abstract}

\keywords{object servoing, differential-drive robots,  motion estimation, asymptotic stabilization}

\section{INTRODUCTION}

Visual servoing is widely used to drive a differential-drive robot from an initial pose to a goal pose  using visual feedback \cite{1chaum2006, 1chaum2007}. The goal pose is fixed and it is predefined by a previously acquired image with a pinhole camera. The differential-drive robot is subject to nonholonomic constraint, which makes the visual servoing task challenging, especially when the final pose is required to be accurate. This is because a nonholonomic system cannot be asymptotically stabilized by any time-invariant continuous state feedback control law due to the well-known Brockett necessary condition \cite{1broc1983}.

Various control methods were developed to asymptotically stabilize the nonholonomic systems such as \cite{1tian2002,1marc2003,1luo2000} in the past decades. The singularity of the nonholonomic systems is deemed undesirable by these methods due to loss of controllability. The singularity set of the differential-drive robot is the set of robot poses with the orientation  being the same as the goal orientation \cite{1tian2002,1marc2003,1luo2000}.  The robot is forced to escape from the singularity set by the controllers in \cite{1tian2002,1marc2003,1luo2000}. Recently, it was found in \cite{1li2018} that the singularity set of the nonholonomic robot includes a subset which is controllable. The subset is determined by the goal pose and it is named as a ``singularity line". On top of switched control \cite{1li2001,1li2005,1lik2003}, a motion controller was designed in \cite{1li2018} for asymptotic stabilization of the differential-drive robot by taking advantage of singular surfaces of the configuration space through the controllable ``singularity line". Hence, the controller in \cite{1li2018} has a {\it unique} feature: it generates {\it straight and smooth motions} (which are singular) naturally when the pose of the robot is in the ``singularity line". The robot also always tries to approach the ``singularity line" under the control of the controller in \cite{1li2018}.

All these controllers assumed that the pose of the robot is already available. This is not true for a visual servoing scheme in which the pose of the robot needs to be estimated.
There are two typical ways to address the visual servoing of mobile robots. Position-based visual servoing (PBVS) of mobile robots was investigated in the literature to reduce the visual servoing task to a control problem in the Cartesian space \cite{1murri2004,1gans2007}. These approaches require 3-D metric information of the features known as a priori, which is further used to reconstruct full-system state for feedback control. An alternative way is to adopt an image-based visual servoing (IBVS) strategy.  Elements of the estimated homography were formulated in \cite{1lopez2007} as the output of the system, based on which a control law was designed by the input–output linearization technique. A novel motion estimation methodology was proposed in \cite{1zhan2011} by using correspondences of three unknown feature points in two images to directly compute the relative pose from images, and a 2-1/2-D visual servoing strategy was then proposed to regulate a nonholonomic mobile robot with an onboard camera to its goal pose. It should be pointed out that the goal pose is fixed in the existing visual servoing systems. However, there are many cases that the goal pose is not fixed but being defined as a relative pose to a movable object. For example, a mobile robot is asked to deliver food to a patient in a movable bed and the bed could be pushed to another pose. Since the goal pose is determined by a given object rather than a previously captured image in \cite{1chaum2006, 1chaum2007}, the new problem is called object servoing. Since the pose of the object can be changed, the existing visual servoing schemes are not applicable to the object servoing. It is thus desired to design an object servoing scheme such that the robot can part itself with a relative pose to a given object regardless of its pose.

A novel object servoing scheme is designed on top of an object servoing friendly controller in this paper. The controller is an improved one of the controller in \cite{1li2018} and it is designed by integrating a fractional-order controller \cite{1chen2009} into the controller in \cite{1li2018}. One uniqueness of the fractional-order controller \cite{1chen2009} is that the system can be asymptotically stabilized within a finite time interval.  The proposed scheme also includes a very simple but efficient way to compute the relative pose for the controller.  The goal pose is defined by the given object in an image which is called a reference image. A deep learning based method is adopted to detect the given object in the reference image \cite{1he2017}. Feature points in the given object are identified and their 3D coordinates are also stored. All these operations are conducted off-line. It is assumed that part of the feature points are visible by the onboard camera of the robot when the robot starts from its initial pose. Each image captured on line is called a query image. The feature points of the given object in the reference image are then matched to those feature points in the query image.  The relative pose of the current pose with respect to the goal one is then estimated for the robot by using a visual based motion estimation algorithm. Besides the proposed pose estimation method, existing pose estimation methods such as \cite{1vidal2018,2li2019} can also be applied to estimate the pose. The complexity of the algorithm in \cite{1vidal2018} is an issue for a real time controller while the accuracy of  the algorithm in \cite{2li2019} needs to be improved.

The proposed scheme also includes an object servoing friendly controller. The relative pose is serve as the input to the improved controller. The corresponding linear velocity and angular velocity are then computed by using the improved controller. The robot moves according to the computed linear velocity and angular velocity. A possible issue for the object servoing is that there might be no common features between the object in some query images and the object in the reference image when the robot moves under the control of the parking controller. This would happen if  the parking controller is not  object servoing friendly. Fortunately, the improved controller  is much more object servoing friendly than the controllers in \cite{1tian2002,1marc2003,1luo2000}. This is because that the given object in the reference image is captured by the onboard camera when the robot is at the goal pose, i.e., the onboard camera is oriented along the ``singularity line". The robot always tries to approach the ``singularity line" under the improved controller. Both simulation and experimental results are used to verify the efficiency of the proposed object servoing scheme. Besides the proposed scheme, an alternative scheme is to first estimate the pose of movable object on-line \cite{1yu2018} and then compute the linear velocity and angular velocity using the improved controller. Compared with such an alternative one, the on-line computational cost of the proposed scheme is much lower. As such, the proposed scheme has a good potential to be adopted for the last mile delivery of differential-drive robots to movable objects.

The rest of this paper is organized as below.  A new problem formulation on object servoing is provided in Section \ref{problemformulation}.  An object servoing scheme is then proposed in Section \ref{objectservoing}. Experimental results are given in Section \ref{experimentalresults} to verify the efficiency of the proposed scheme. Finally, conclusion remarks are given in Section \ref{conclusion}.

\section{Problem Formulation on Object Servoing}
\label{problemformulation}

\begin{figure}[htb]
	\centering{
		\includegraphics[width=2.86in]{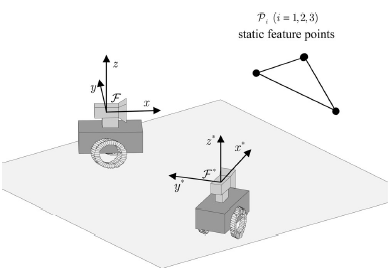}}
	\caption{the camera coordinate system for a real robot platform \cite{1zhan2011}.}
	\label{real_robot}
	\vspace{-0.3cm}
\end{figure}

A differential-drive mobile robot that is equipped with an onboard camera is considered in this paper and the robot is shown in Fig. \ref{real_robot}.
The  motion of the robot is  described in terms of a rotation matrix $R\in \mathbb{R}^{3\times 3}$ and a translation vector $T\in \mathbb{R}^3$ as
\begin{equation}
	R=\left[\begin{array}{lll}
		\cos\theta&-\sin\theta & 0\\
		\sin\theta & \cos\theta & 0\\
		0 & 0 & 1\\
	\end{array}
	\right]\; ;\; T=\left[\begin{array}{l}
		t_x\\t_y\\0\\
	\end{array}
	\right],
\end{equation}
where $\theta\in \mathbb{R}$ is the rotational angle around the z-axis.

Let $F$ be an orthogonal coordinate system which is attached to the on-board camera, with its origin  being located at the camera center. The $z$-axis is supposed to pass through
the midpoint of the wheels axis, and it is orthogonal to the motion plane of the mobile robot. The $x$-axis is along the optical axis, and it is aligned with the front of the robot. In addition, one more orthogonal coordinate system $F^*$ is adopted to represent the goal pose of the camera. For simplicity, $F$ and $F^*$ are called a query image and a reference image, respectively. A deep learning based method was used off-line to detect the given object in the reference image \cite{1he2017}. The 3D coordinates of the feature points in the reference image are stored. All these operations are conducted off-line. The goal pose is not fixed but a relative pose to the given movable object such as a bed or a chair. Such a problem is named as object servoing.

Instead of selecting feature points arbitrarily from the reference image and query image as in \cite{1zhan2011},  feature points are selected from the given object.  Let $m_i = \left[\begin{array}{ll} u_i & v_i\end{array}\right]^T$ be the $i$th feature point of the movable object in the current frame  and  $P_i = \left[\begin{array}{lll} X_i &
	Y_i & Z_i\end{array}\right]^T$ be the coordinate of the corresponding pixel in the current frame.
The relationship between $m_i$ and  $P_i$
represented as
\begin{equation}
	\left[\begin{array}{l}
		u_i\\
		v_i\\
	\end{array}
	\right]=\pi(P_i)
	=\left[\begin{array}{l}
		f_x\frac{Y_i}{X_i}+c_x\\
		f_y\frac{Z_i}{X_i}+c_y\\
	\end{array}
	\right],
\end{equation}
where $f_x$  and $f_y$  are the focal lengths of the camera in pixels, and $c_x$ and $c_y$  are the coordinates of the principle point in pixels.
For simplicity, a vector $\left[\begin{array}{ll} x_i & y_i\end{array}\right]^T$ is defined as
\begin{equation}
	\label{feature1}
	\left[\begin{array}{l}
		x_i\\
		y_i\\
	\end{array}
	\right]\doteq \left[\begin{array}{l}
		\frac{Y_i}{X_i}\\
		\frac{Z_i}{X_i}\\
	\end{array}
	\right]
	=\left[\begin{array}{l}
		\frac{u_i-c_x}{f_x}\\
		\frac{v_i-c_y}{f_y}\\
	\end{array}
	\right].
\end{equation}

Similarly, let $m_i^* = \left[\begin{array}{ll} u_i^* & v_i^*\end{array}\right]^T$ be the $i$th feature point of the movable object in the goal frame,  $P_i^*(= \left[\begin{array}{lll} X_i^* &
	Y_i^* & Z_i^*\end{array}\right]^T)$ be the corresponding coordinate  in the goal frame. A vector $\left[\begin{array}{ll} x_i^* & y_i^*\end{array}\right]^T$ is defined as
\begin{equation}
	\label{xiyi2d}
	\left[\begin{array}{l}
		x_i^*\\
		y_i^*\\
	\end{array}
	\right]\doteq \left[\begin{array}{l}
		\frac{Y_i^*}{X_i^*}\\
		\frac{Z_i^*}{X_i^*}\\
	\end{array}
	\right]
	=\left[\begin{array}{l}
		\frac{u_i^*-c_x}{f_x}\\
		\frac{v_i^*-c_y}{f_y}\\
	\end{array}
	\right].
\end{equation}

To simplify the on-line computation, it is assumed that  $X_i^*$s are available. They can be determined using any existing method off-line.
The relationship between $P_i$ and $P_i^*$ is given as
\begin{equation}
	\label{eqeq1}
	P_i=RP_i^*+T.
\end{equation}

Similar to \cite{1zhan2011}, define $x_e$ and $y_e$  as
\begin{equation}
	\label{xeye2d}
	\left[\begin{array}{l}
		x_e\\
		y_e\\
	\end{array}
	\right]=\left[\begin{array}{l}
		\frac{1}{y_i}\\
		\frac{x_i}{y_i}\\
	\end{array}
	\right]-\left[\begin{array}{ll}
		\sin\theta & \cos\theta\\
		\cos\theta & -\sin\theta\\
	\end{array}
	\right]\left[\begin{array}{l}
		\frac{x_i^*}{y_i^*}\\
		\frac{1}{y_i^*}\\
	\end{array}
	\right],
\end{equation}
and $\theta_e$ as $-\theta$. It can be derived that
\begin{equation}
	\label{xedot}
	\left[\begin{array}{l}
		\dot{x}_{e}\\
		\dot{y}_{e}\\
		\dot{\theta}_{e}\\
	\end{array}\right]
	=\left[\begin{array}{lll}
		0 & \omega & 0\\
		-\omega & 0 & 0\\
		0 & 0 & 0\\
	\end{array}
	\right]\left[\begin{array}{l}
		x_{e}\\
		y_{e}\\
		\theta_{e}\\
	\end{array}\right]-
	\left[\begin{array}{l}
		\frac{v}{Z_i}\\
		0\\
		\omega\\
	\end{array}\right],
\end{equation}
where $v$ and $\omega$ are the linear velocity and angular velocity of the on-board camera.

Defining a new set of states  as
\begin{align}
	\left\{\begin{array}{l}
		z_0 = -\theta_e=\theta,\\
		z=\left[\begin{array}{ll} z_1 & z_2
		\end{array}
		\right]^T
		= \left[\begin{array}{ll}  y_e & -x_e
		\end{array}
		\right]^T
	\end{array}
	\right.,
\end{align}
and  a new set of control inputs as \cite{1orio2002}
\begin{eqnarray}
	\left[\begin{array}{l}u_0\\
		u_1\\
	\end{array}
	\right] = \left[\begin{array}{l}\omega\\
		\frac{v}{Z_i}-z_1 u_0\\
	\end{array}
	\right],
\end{eqnarray}
it follows that
\begin{eqnarray}
	\label{state}
	\left[\begin{array}{l}\dot{z}_0\\
		\dot{z}\\
	\end{array}
	\right]
	=\left[\begin{array}{l} u_0\\
		u_0Az+Bu_1\\
	\end{array}
	\right],
\end{eqnarray}
where the matrix $A$ and the vector $B$ are
\begin{align}
	A&=\left[\begin{array}{ll}
		0 & 1\\
		0 & 0\\
	\end{array}
	\right]\; ;\; B =\left[\begin{array}{ll} 0 & 1\end{array}
	\right]^T.
\end{align}

The object servoing problem is formulated as follows:

{\it Regulate the differential-drive robot to its desired pose by image feedback such that the goal poses of the robot and the movable object satisfy the predefined relative relationship by $F^*$ regardless of the pose of the movable object, i.e., both $z_0$ and $z$ approach zeros. The linear velocity and angular velocity of the differential-drive robot are zeros when the relative relationship is satisfied, i.e., both $u_0$ and $u_1$ are zeros when $z_0$ and $z$ are zeros.}

The object servoing is different from the existing visual servoing in the sense that the goal pose is fixed in the visual servoing while it is not fixed in the object servoing. The goal pose in the object servoing is determined by the pose of a given object which could be changed.

%

%
%
%
%
%
%

\section{The Proposed Object Servoing}
\label{objectservoing}
An object servoing scheme will be designed in this section. The proposed scheme consists of  an object servoing friendly parking controller  and a simple visual motion estimation algorithm. The controller is obtained by integrating a fractional-order controller into the switched controller in \cite{1li2018}.

\subsection{An Object Servoing Friendly Parking Controller}

Similar to \cite{1li2018}, it can be shown that the system (\ref{state}) is asymptotically stabilized by the improved parking controller in {\it Algorithm \ref{algo2}}.
\begin{algorithm}
	{
		\caption{An improved asymptotic stabilizer  of the error system (\ref{state})}
		
		\begin{enumerate}[itemindent=1em]
			\item[Inputs]: $\kappa_0>0$, $\kappa_2>0$, $\epsilon>1$, $\xi>0$, and $\delta>0$ \cite{1marc2003}.
			\item[Step 1.] $\gamma = \kappa_0\epsilon+\xi$, $\zeta= 2\gamma+\kappa_0$, and\vspace{2mm}  $\kappa_1=\frac{2\gamma+\sqrt{\kappa_0^2+6\kappa_0\zeta+\zeta^2}+1}{4}$.
			\item[Step 2.] Compute three constants $P_i(i=1,2,3)$ as
			\begin{eqnarray}
				\left[\begin{array}{l}
					P_3\\
					P_2\\
					P_1\\
				\end{array}
				\right] = \left[\begin{array}{l}\frac{\kappa_0+3\zeta+\sqrt{\kappa_0^2+6\kappa_0\zeta+\zeta^2}}{4}\\
					P_3^2-\frac{\kappa_0+\zeta}{2}P_3\\
					\frac{2P_2^2}{\zeta}\\
				\end{array}
				\right].
			\end{eqnarray}
			\item[Step 3.] Define the controller $u_0$ as \cite{1li2018,1chen2009}
			\begin{eqnarray}
				\label{u0}
				&&\hspace{-23mm}u_0=\left\{\begin{array}{ll}
					-z_0^{\frac{1}{3}}; &\mbox{if~}z_1=z_2=0\\
					-\kappa_0 z_0; &\mbox{otherwise if~}\left[\begin{array}{lll}z_0 & z_1 & z_2
					\end{array}
					\right]\in \Gamma\\
					\frac{-\kappa_1z_1}{\psi(z_2)}; &\mbox{otherwise}\\
				\end{array}
				\right.,\\
				\label{u0u0}
				&&\hspace{-23mm}\psi(z_2)=\left\{\begin{array}{ll}
					z_2; &\mbox{if~}z_2\neq 0\\
					\mbox{sign}(z_0z_1); &\mbox{otherwise}\\
				\end{array}
				\right..
			\end{eqnarray}

			\item[Step 4.] Define the controller $u_1$ as \cite{1li2018,1chen2009}
			\begin{equation}
				\label{u1}
				\hspace{-5mm}
				u_1=\left\{\begin{array}{ll}
					-z_2^{\frac{1}{3}}; &\mbox{if~}z_1=z_0=0\\
					-\kappa_2z_2; &\mbox{otherwise if~}u_0=0\\
					-\frac{P_2}{u_0}z_1-P_3z_2; &\mbox{otherwise}\\
				\end{array}
				\right..
			\end{equation}
		\end{enumerate}
		\label{algo2}
	}
\end{algorithm}

In  {\it Algorithm \ref{algo2}},  $P_i(i=1,2,3)$ are obtained by solving the following Riccati equation \cite{1marc2003}:
\begin{equation}
	\label{matrixP}
	A^TP+PA-2PBB^TP+(2\gamma+\kappa_0)P+\kappa_0 LPL =0,
\end{equation}
where $\kappa_0$ is a positive constant.  The value of $\gamma$ is $(\kappa_0\epsilon+\xi)$ with $\epsilon(>1)$ and $\xi$ being two positive constants. Two matrices $P$ and $L$  are defined as
\begin{eqnarray}
	&&\hspace{-11mm} P=\left[\begin{array}{ll}
		P_1 & P_2\\
		P_2 & P_3\\
	\end{array}
	\right]\; ;\; L=\left[\begin{array}{ll}
		0 & 0\\
		0 & 1\\
	\end{array}
	\right].
\end{eqnarray}

The invariant set $\Gamma$ is defined  as follows \cite{1li2018}:
\begin{equation}
	\label{Gamma} \{\left[\begin{array}{lll}z_0 & z^T
	\end{array}
	\right]|V(z_0, z)\,\textless \,\delta(\kappa_0 z_0)^{2\epsilon}\mbox{~or~}|z_0|+|z_1|=0\},
\end{equation}
where $\delta$ is a positive constant, and the function $V(z_0,z)$ is defined as
\begin{eqnarray}
	V(z_0, z) = P_1z_1^2-2P_2\kappa_0 z_0z_1z_2+P_3\kappa_0^2z_0^2z_2^2.
\end{eqnarray}

Without loss of generality, it is assumed that there are common feature points between the movable object in the first query image and  the object in the reference image. A parking controller is object servoing friendly if there usually exist common feature points between the movable object in the subsequent query image and  the object in the reference image when the robot is moved under the controller.  The controller in {\it Algorithm \ref{algo2}} is analyzed as below.

The invariant set $\Gamma$ includes a set $S_{s,c}$ which is defined as
\begin{equation}
	\label{singsta}
	S_{s,c}=\{(x_e, y_e, \theta_e)| \theta_e = 0, y_e =0\},
\end{equation}
and the set $S_{s,c}$ is actually part of the ``singularity line" which is determined by the goal pose. It can be easily verified that the set $S_{s,c}$ is a subset of the following set $S_s$:
\begin{equation}
	S_s=\{(x_e, y_e, \theta_e)|\theta_e = 0\},
\end{equation}
It is believed in \cite{1marc2003,1luo2000,1zhan2011} that the system is uncontrollable in the  singularity set $S_s$ and it is challenging to design a controller for asymptotic stabilization of a differential-drive robot when the pose of the robot is in the singular set. Therefore, the  set  $S_{s,c}$ is being avoided or escaped  by the existing parking control algorithms \cite{1marc2003,1luo2000,1zhan2011}. On the other hand, the object is well observed if the pose of the robot is in the set $S_{s,c}$.

Defining a function of $z_1$ as
\begin{equation}
	\varphi(z_1) = \frac{1}{2}z_1^2,
\end{equation}
it can be verified that
\begin{equation}
	\frac{d\varphi(z_1)}{dt} = \left\{\begin{array}{ll}
		-\kappa_1 z_1^2; &\mbox{if~}z_2\neq 0\\
		0; &\mbox{otherwise}\\
	\end{array}
	\right..
\end{equation}

This implies that the set $S_{s,c}$ is approached by the controller in {\it Algorithm \ref{algo2}} if the state  $\left[\begin{array}{lll}z_0 & z_1 & z_2 \end{array}\right]^T$ is not in the set $\Gamma$. Therefore, the controller in {\it Algorithm \ref{algo2}} is more object servoing friendly than the controller in \cite{1marc2003,1luo2000,1zhan2011}.

\subsection{A Simple Visual Motion Estimation Algorithm}

$z_i(i=0,1,2)$ are used by the controller in {\it Algorithm \ref{algo2}}. According to the equations (\ref{xiyi2d})-(\ref{xeye2d}), both $R$ and $T$ will be required to compute $z_i(i=0,1,2)$ for each query image. Once a query image is captured on-line, the feature points of the given object in the reference image will match those feature points in the query image. All matched pairs will be utilized to estimate $R$ and $T$ for the query image. It should be pointed out that the exposures of the query image and the reference image could be differently \cite{1j2011,1G2013,lFkou2017}. Their matching could be addressed by using the mapping method in \cite{1jjiang2020}.

$\sin\theta$, $\cos\theta$, $t_x$ and $t_y$ are computed by using  matched feature points of the given object between each query image and the reference image.  For simplicity,  $\sin\theta$ and $\cos\theta$ are first computed. $t_x$ and $t_y$ are then computed separately.

Considering two pairs of matched feature points $(P_i, P_i^*)$ and $(P_j, P_j^*)$, it can be derived from Equations (\ref{feature1})-(\ref{eqeq1}) that
\begin{equation}
	\label{pipi}
	\left[\begin{array}{l}
		x_l\\
		y_l\\
	\end{array}
	\right]=
	\left[\begin{array}{l}
		\frac{X_l^*(x_l^*\cos\theta+sin\theta)+t_y}{X_l^*(cos\theta-x_l^*\sin\theta)+t_x}\\
		\frac{X_l^*y_l^*}{X_l^*(cos\theta-x_l^*\sin\theta)+t_x}\\
	\end{array}
	\right]\; ;\;l\in\{i,j\}.
\end{equation}

It follows that \cite{1zhan2011}
\begin{equation}
	a_{ij}\sin\theta+b_{ij}\cos\theta+c_{ij}=0,
\end{equation}
where $\left[\begin{array}{lll} a_{ij} & b_{ij} & c_{ij}\end{array}\right]^T$  is computed by
\begin{equation}
	\left[\begin{array}{l}
		a_{ij}\\
		b_{ij}\\
		c_{ij}\\
	\end{array}
	\right]= \left[\begin{array}{l}
		\frac{y_i^*}{y_i}(x_ix_j^*+1)-\frac{y_j^*}{y_j}(x_i^*x_j+1)\\
		\frac{y_i^*}{y_i}(x_j^*-x_i)-\frac{y_j^*}{y_j}(x_i^*-x_j)\\
		\frac{y_i^*y_j^*}{y_iy_j}(x_i-x_j)+(x_i^*-x_j^*)\\
	\end{array}
	\right].
\end{equation}

The optimal values of $\sin\theta$ and $\cos\theta$ can be obtained by solving the following optimization problem:
\begin{equation}
	\arg\min_{\sin\theta,\cos\theta}\{\sum_{i,j}(a_{ij}\sin\theta+b_{ij}\cos\theta+c_{ij})^2\},
\end{equation}
s.t.
\begin{equation}
	\label{sincos1}
	\sin^2\theta+\cos^2\theta=1.
\end{equation}

Let the cost function of the above optimization problem be denoted as $\Phi(\sin\theta,\cos\theta)$. The function $\Phi(\sin\theta,\cos\theta)$ is computed as
\begin{equation}
	\Phi=\sum_{i,j}(a_{ij}\sin\theta+b_{ij}\cos\theta+c_{ij})^2+\lambda(\sin^2\theta+\cos^2\theta-1).
\end{equation}
It can be derived from the partial derivatives that the optimal values of  $\sin\theta$ and $\cos\theta$ are
\begin{equation}
	\label{sincos2}
	\left[\begin{array}{l}
		\sin\theta\\
		\cos\theta\\
	\end{array}
	\right]= \left[\begin{array}{l}
		\frac{(\tilde{a}_3+\lambda)\tilde{b}_1-\tilde{a}_2\tilde{b}_2}{(\tilde{a}_1+\lambda)(\tilde{a}_3+\lambda)-\tilde{a}_2^2}\\
		\frac{(\tilde{a}_1+\lambda)\tilde{b}_2-\tilde{a}_2\tilde{b}_1}{(\tilde{a}_1+\lambda)(\tilde{a}_3+\lambda)-\tilde{a}_2^2}\\
	\end{array}
	\right],
\end{equation}
where  $\left[\begin{array}{lll} \tilde{a}_1 & \tilde{a}_2 & \tilde{a}_3\end{array}\right]^T$  and $\left[\begin{array}{ll} \tilde{b}_1 & \tilde{b}_2\end{array}\right]^T$ are computed by
\begin{align}
	&\left[\begin{array}{l}
		\tilde{a}_1\\
		\tilde{a}_2\\
		\tilde{a}_3\\
	\end{array}
	\right]= \left[\begin{array}{l}
		{\displaystyle \sum_{i,j}}a_{ij}^2\\
		{\displaystyle \sum_{i,j}}a_{ij}b_{ij}\\
		{\displaystyle \sum_{i,j}}b_{ij}^2\\
	\end{array}
	\right],\\
	&\left[\begin{array}{l}
		\tilde{b}_1\\
		\tilde{b}_2\\
	\end{array}
	\right]= \left[\begin{array}{l}
		-{\displaystyle \sum_{i,j}}a_{ij}c_{ij}\\
		-{\displaystyle \sum_{i,j}}b_{ij}c_{ij}\\
	\end{array}
	\right].
\end{align}

Combining Equations (\ref{sincos1}) and (\ref{sincos2}), it can be obtained that
\begin{equation}
	\label{lambda}
	\lambda^4+2c_1\lambda^3+c_2\lambda^2+2c_3\lambda+c_4=0,
\end{equation}
where $\left[\begin{array}{llll}c_1 & c_2 & c_3 & c_4\end{array}\right]^T$ is computed as
{\small \begin{equation} \left[\begin{array}{l}
			c_1\\
			c_2\\
			c_3\\
			c_4\\
		\end{array}
		\right]=
		\left[\begin{array}{l}
			\tilde{a}_1+\tilde{a}_3\\
			(\tilde{a}_1+\tilde{a}_3)^2-\tilde{b}_1^2-\tilde{b}_2^2+2(\tilde{a}_1\tilde{a}_3-\tilde{a}_2^2)\\
			(\tilde{a}_1+\tilde{a}_3)(\tilde{a}_1\tilde{a}_3-\tilde{a}_2^2)+2\tilde{a}_2\tilde{b}_1\tilde{b}_2-\tilde{a}_3\tilde{b}_1^2-\tilde{a}_1\tilde{b}_2^2\\
			(\tilde{a}_1\tilde{a}_3-\tilde{a}_2^2)^2-(\tilde{a}_3\tilde{b}_1-\tilde{a}_2\tilde{b}_2)^2-(\tilde{a}_1\tilde{b}_2-\tilde{a}_2\tilde{b}_1)^2\\
		\end{array}
		\right].
\end{equation}}

Once $\lambda$ is obtained from Equation (\ref{lambda}),  $\sin\theta$ and $\cos\theta$ can be computed using Equation (\ref{sincos2}).

When  $t_x$ and $t_y$ are computed, the 3D information of feature points in one of the reference and query images  is required. To reduce the on-line complexity, the reference image is supposed to include the 3D information of the feature points while the query image only includes the 2D information of the feature points.

Using Equation (\ref{pipi}), $t_x$ and $t_y$ are  obtained via solving the following optimization problem:
\begin{equation}
	\arg\min_{t_x,t_y}\{\sum_{i}[(d_i-t_x)^2+(e_i+x_it_x-t_y)^2]\},
\end{equation}
where  $d_i$ and $e_i$ are
\begin{equation}
	\left[\begin{array}{l}
		d_i\\
		e_i\\
	\end{array}
	\right]=
	\left[\begin{array}{l}
		X_i^*(\frac{y_i^*}{y_i}-(\cos\theta-x_i^*\sin\theta))\\
		X_i^*((x_i-x_i^*)\cos\theta-(x_ix_i^*+1)\sin\theta)\\
	\end{array}
	\right].
\end{equation}

Their optimal values are computed as
\begin{equation}
	\left[\begin{array}{l}
		t_x\\
		t_y\\
	\end{array}
	\right]= \left[\begin{array}{l}
		\frac{N\sum_i(d_i-x_ie_i)+\sum_ix_i\sum_ie_i}{N\sum_i(1+x_i^2)-(\sum_ix_i)^2}\\
		\frac{1}{N}(\sum_ix_i t_x+\sum_i e_i)\\
	\end{array}
	\right].
\end{equation}

Compared with a possible alternative by estimating the pose of the movable object on-line \cite{1yu2018}, the on-line computational cost of the proposed motion estimation method is much lower.

\section{Simulation  Results}
\label{experimentalresults}
In this section, simulation results are  provided to illustrate the proposed controller  is more object servoing friendly. The simulation results for the controller in \cite{1marc2003} are also provided to compare the two controllers. The parameters of the existing controller are selected according to the recommendation in the original paper \cite{1marc2003}.  The controller parameters are summarised in Table \ref{parameter}. Four sets of initial and target poses are listed in Table  \ref{posecase}.

 \begin{figure}[h]
	\centering{
		\subfigure[]
		{\includegraphics[width=1.5in]{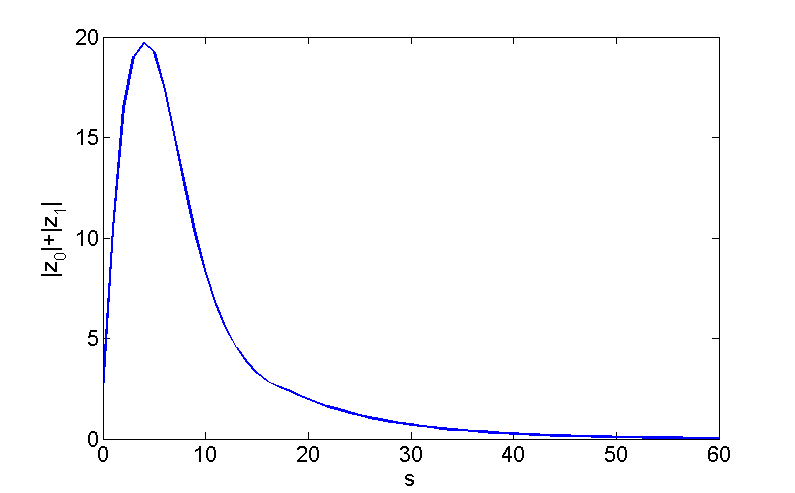}}
		\subfigure[]
		{\includegraphics[width=1.5in]{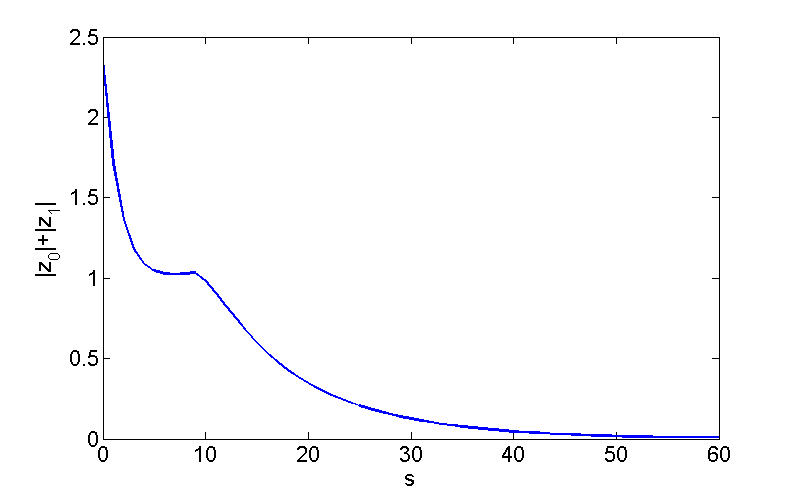}}
		\subfigure[]
		{\includegraphics[width=1.5in]{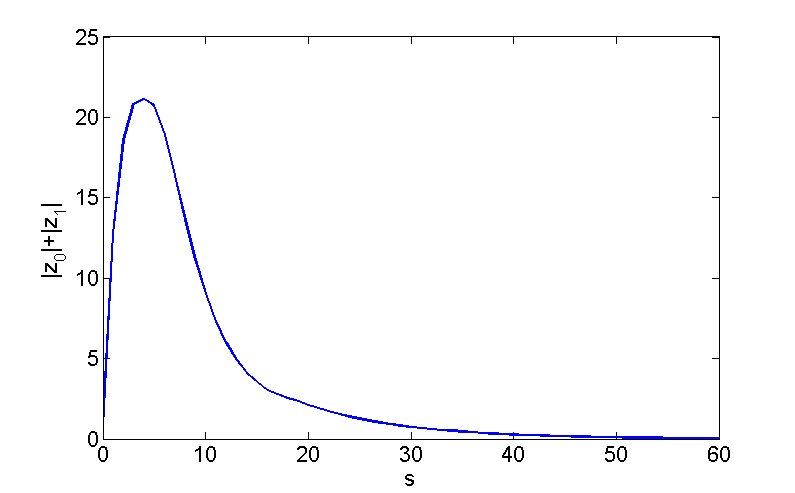}}
		\subfigure[]
		{\includegraphics[width=1.5in]{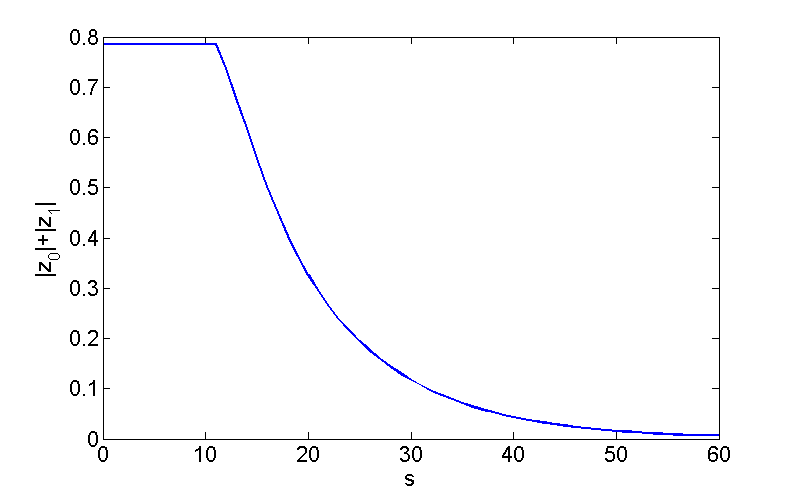}}}
	
	\caption{Curves of $|z_0|+|z_1|$. (a) and (c) by the controller in \cite{1marc2003}; (b) and (d) by the controller in this paper.}
	\label{figcontroloutput}
	\vspace{-0.1cm}
\end{figure}

\begin{table}[!h]
	\caption{PARAMETER  FOR  CONTROLLERS}
	\label{table}
	\begin{center}
		\tabcolsep 0.06in
		\renewcommand{\arraystretch}{1.5}
		\begin{tabular}{|c|c|c|c|c|c|c|c|}
			\hline
			Controller & $\beta$ & $\kappa_0$ & $\alpha$ &  $\kappa_2$ & $\epsilon$ & $\xi$ & $\delta$\\
			\hline
			Controller in \cite{1marc2003} & 2& 0.1 & $\kappa_0$ & N.A. & 2.05 & $\frac{1}{1024}$ & 25\\\hline
			Proposed controller  & N.A.&0.1 & $\kappa_0$ &  $\frac{1}{4}$ & 2.25 & $\frac{1}{1024}$ & 25\\\hline
		\end{tabular}
	\end{center}
	\label{parameter}
\end{table}

\begin{figure*}[htb]
	\centering
	\subfigure[]
	{\includegraphics[width=1.65in]{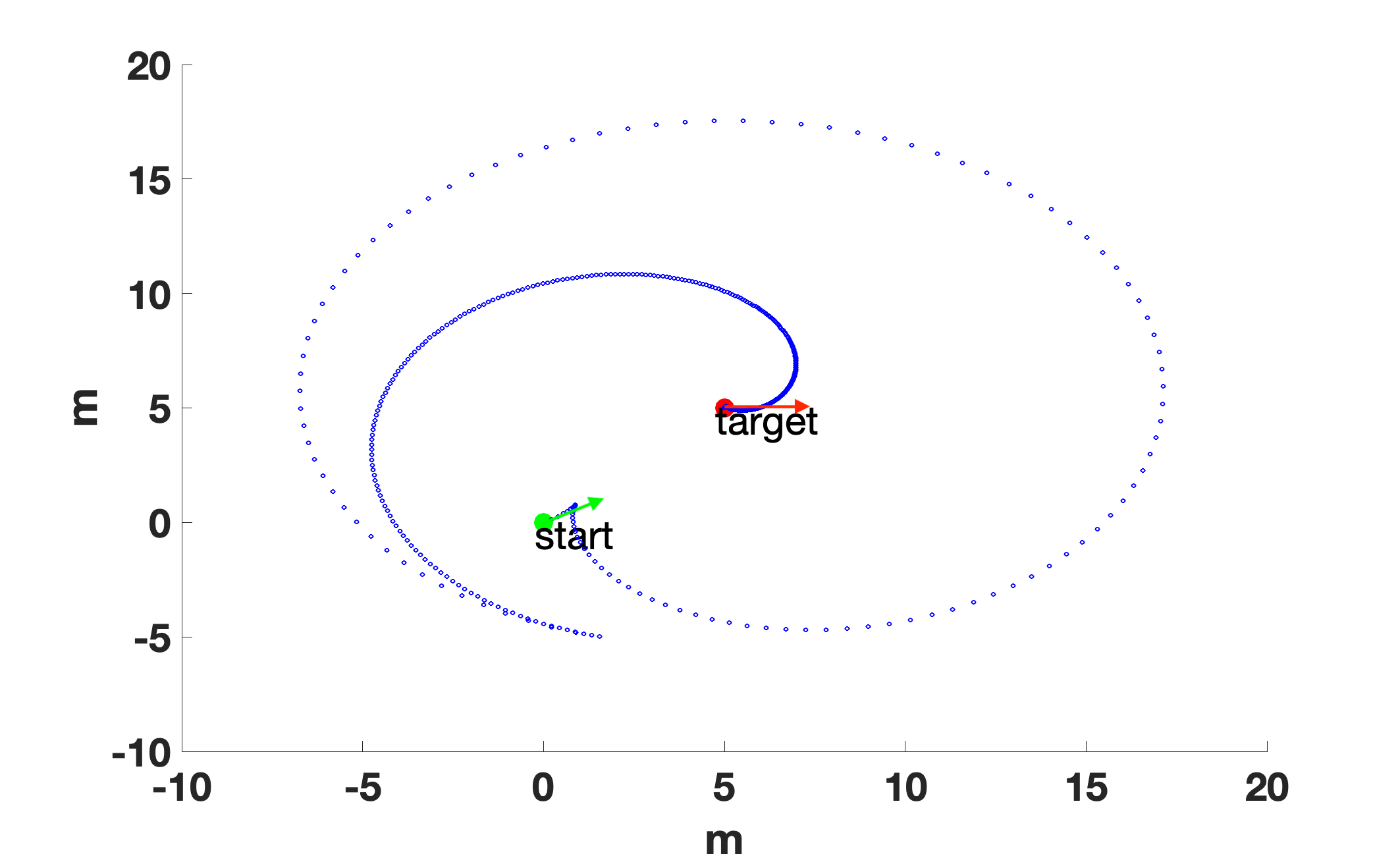}}
	\subfigure[]
	{\includegraphics[width=1.65in]{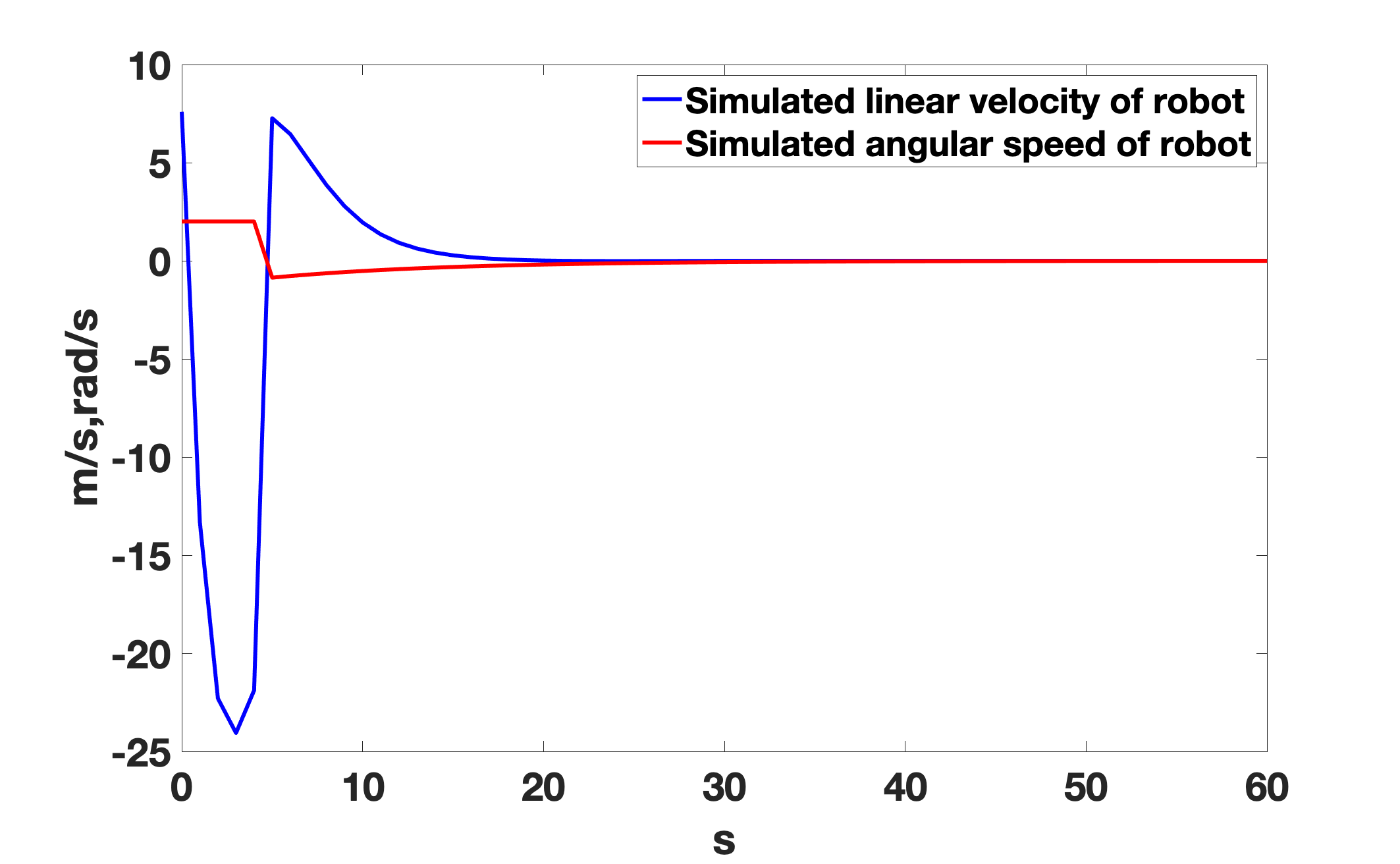}}
	\subfigure[]
	{\includegraphics[width=1.65in]{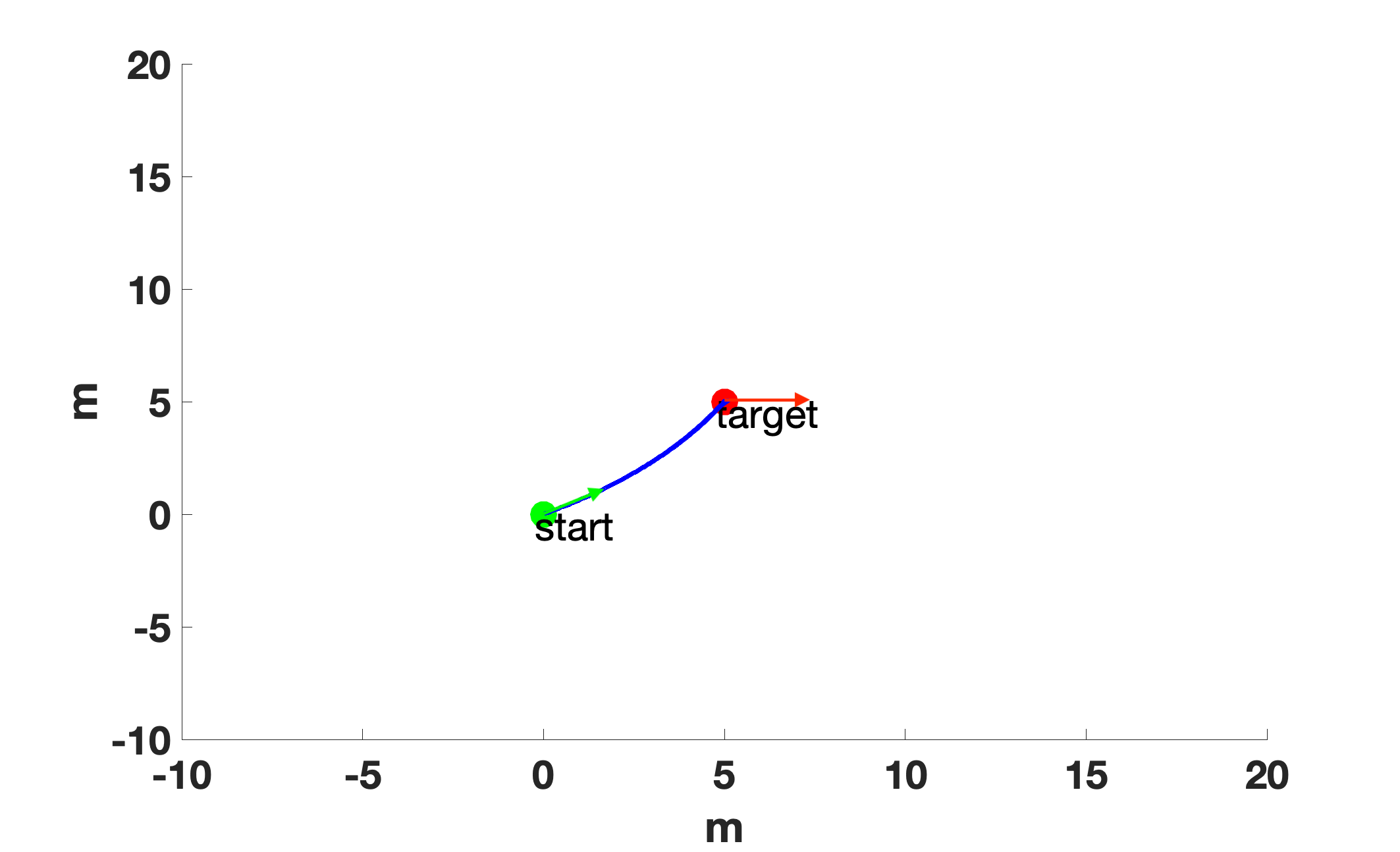}}
	\subfigure[]
	{\includegraphics[width=1.65in]{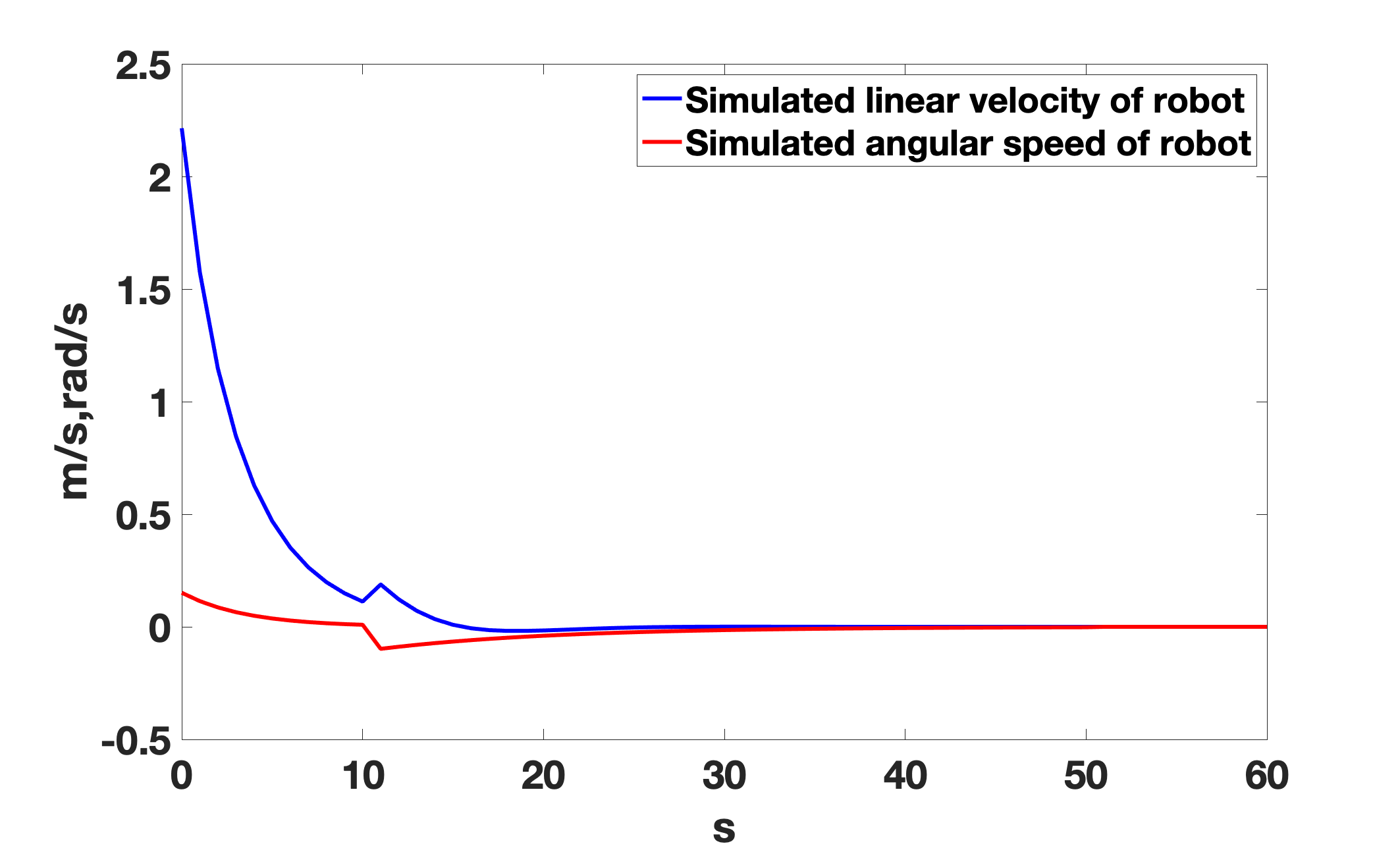}}
	\subfigure[]
	{\includegraphics[width=1.65in]{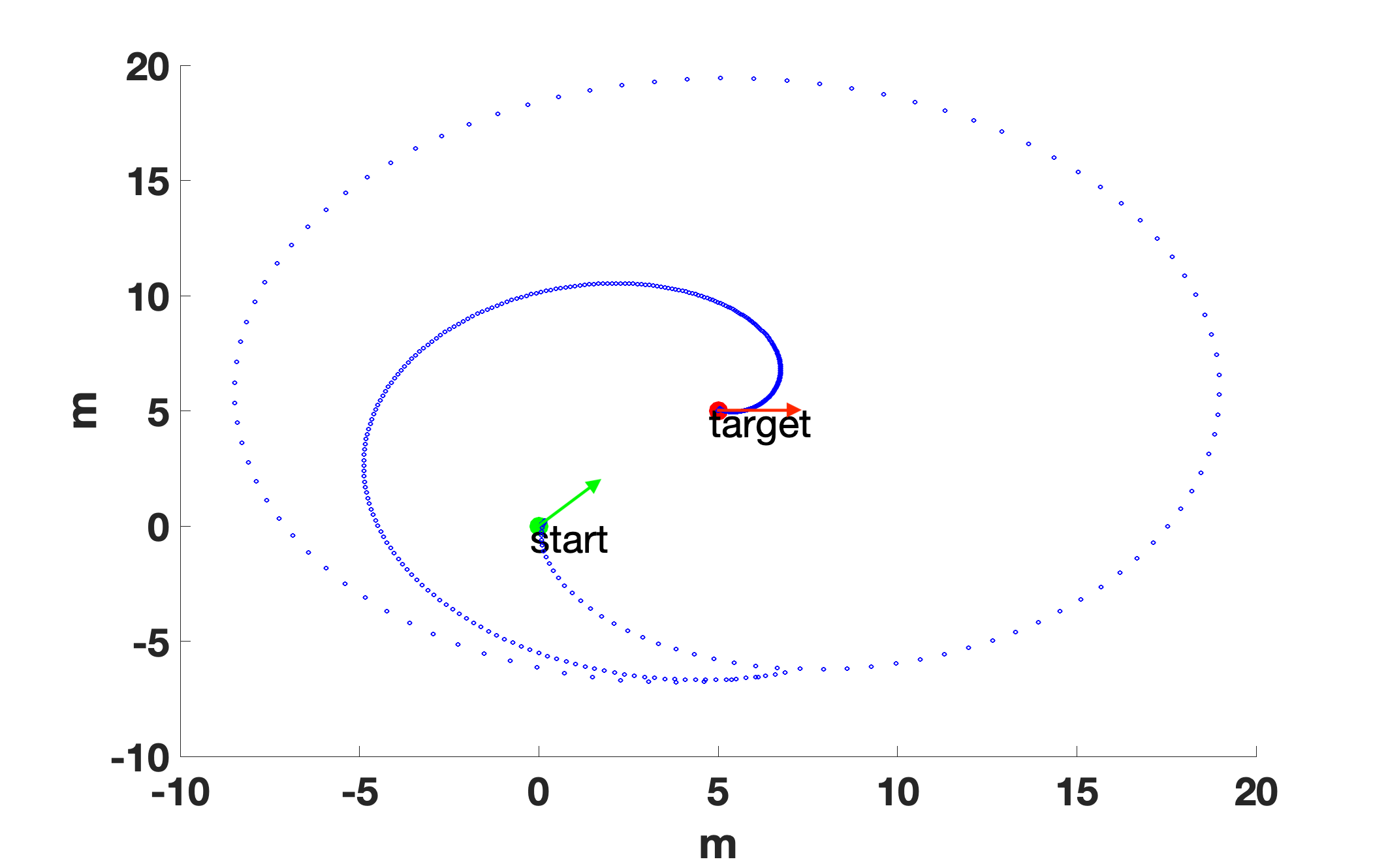}}
	\subfigure[]
	{\includegraphics[width=1.65in]{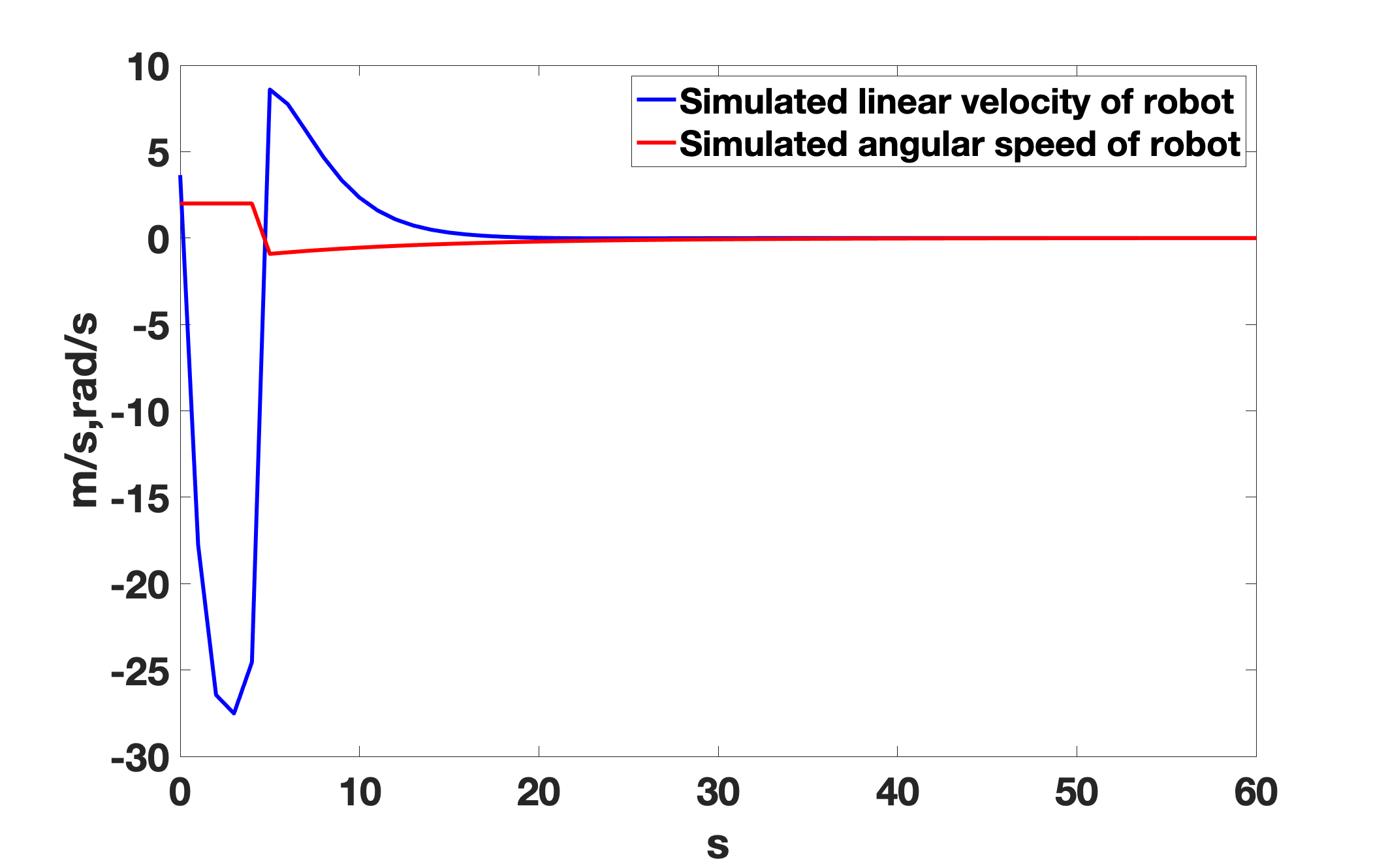}}
	\subfigure[]
	{\includegraphics[width=1.65in]{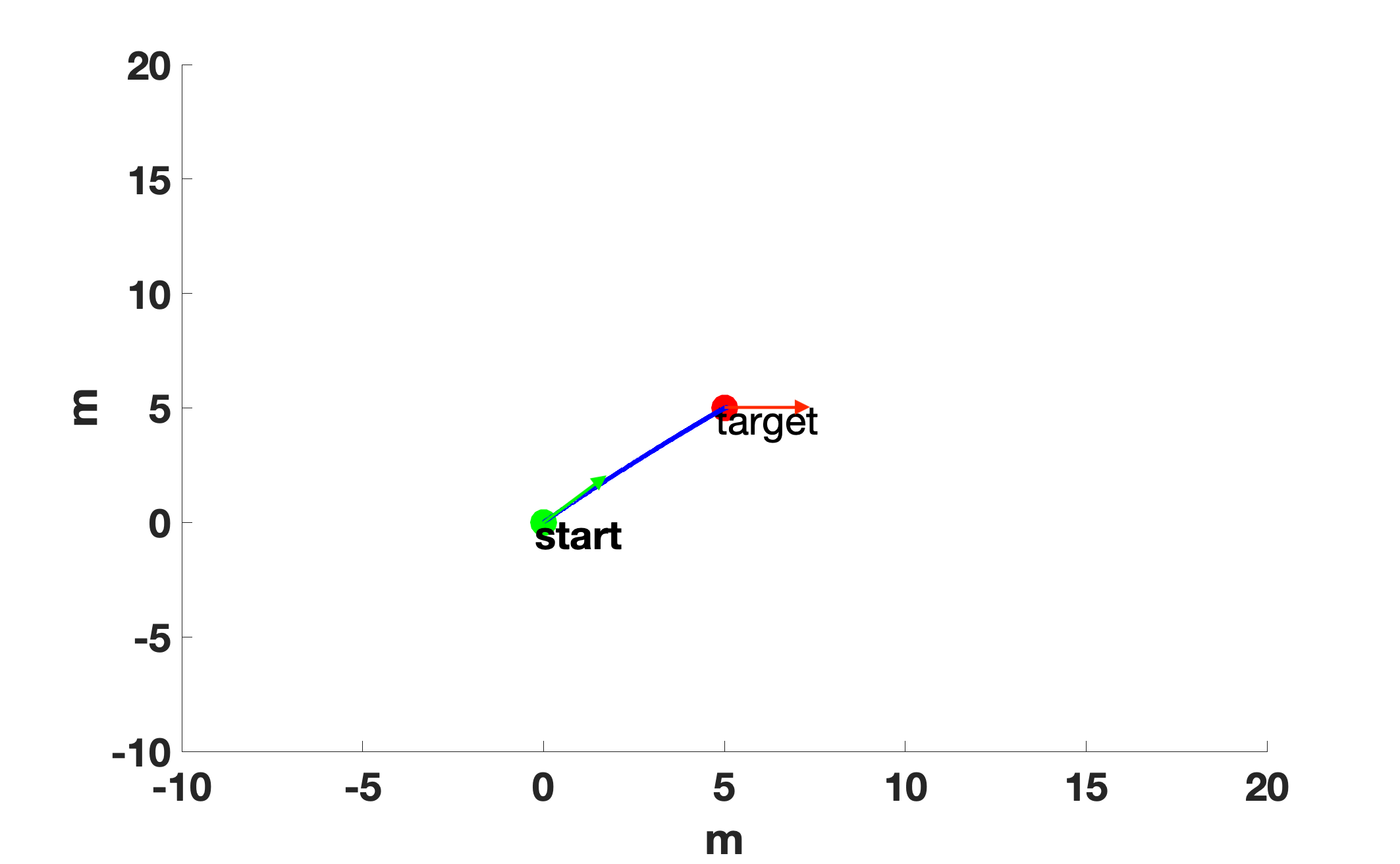}}
	\subfigure[]
	{\includegraphics[width=1.65in]{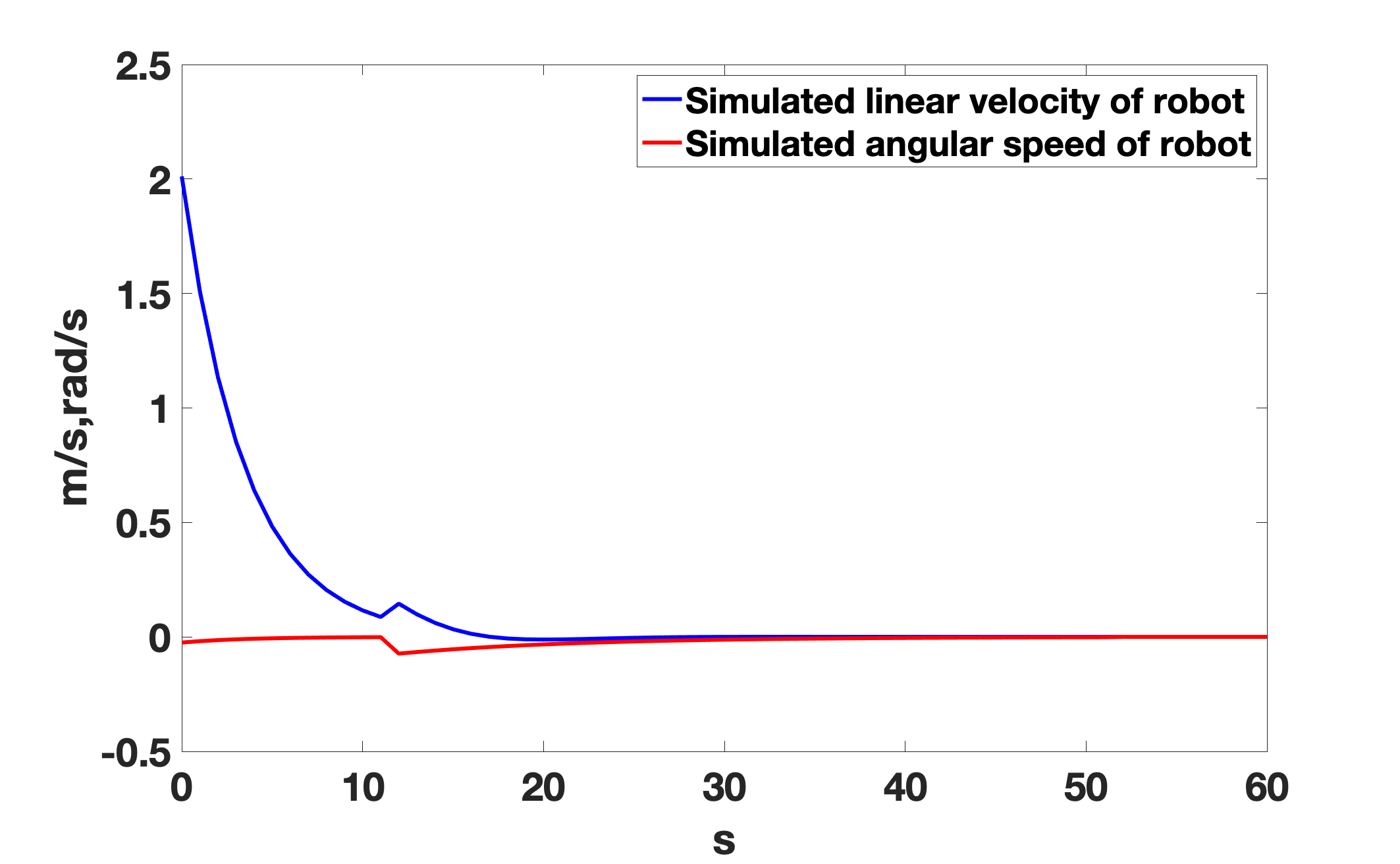}}
	
	\caption{Simulation trajectories (left) and control outputs (right) using the proposed controller and the controller in \cite{1marc2003} without kinematic constraints. (a)
		and (b): Trajectory and output by the controller in \cite{1marc2003} for Case 1; (c) and (d): Trajectory and output by the proposed controller for Case 1; (e) and (f): Trajectory and output by the controller in \cite{1marc2003} for Case 2; (g) and (h): Trajectory and output by the proposed controller for Case 2.}
	\label{figwithoutkc}
	\vspace{-0.3cm}
\end{figure*}

\begin{figure*}[htb]
	\centering
	\subfigure[]
	{\includegraphics[width=1.65in]{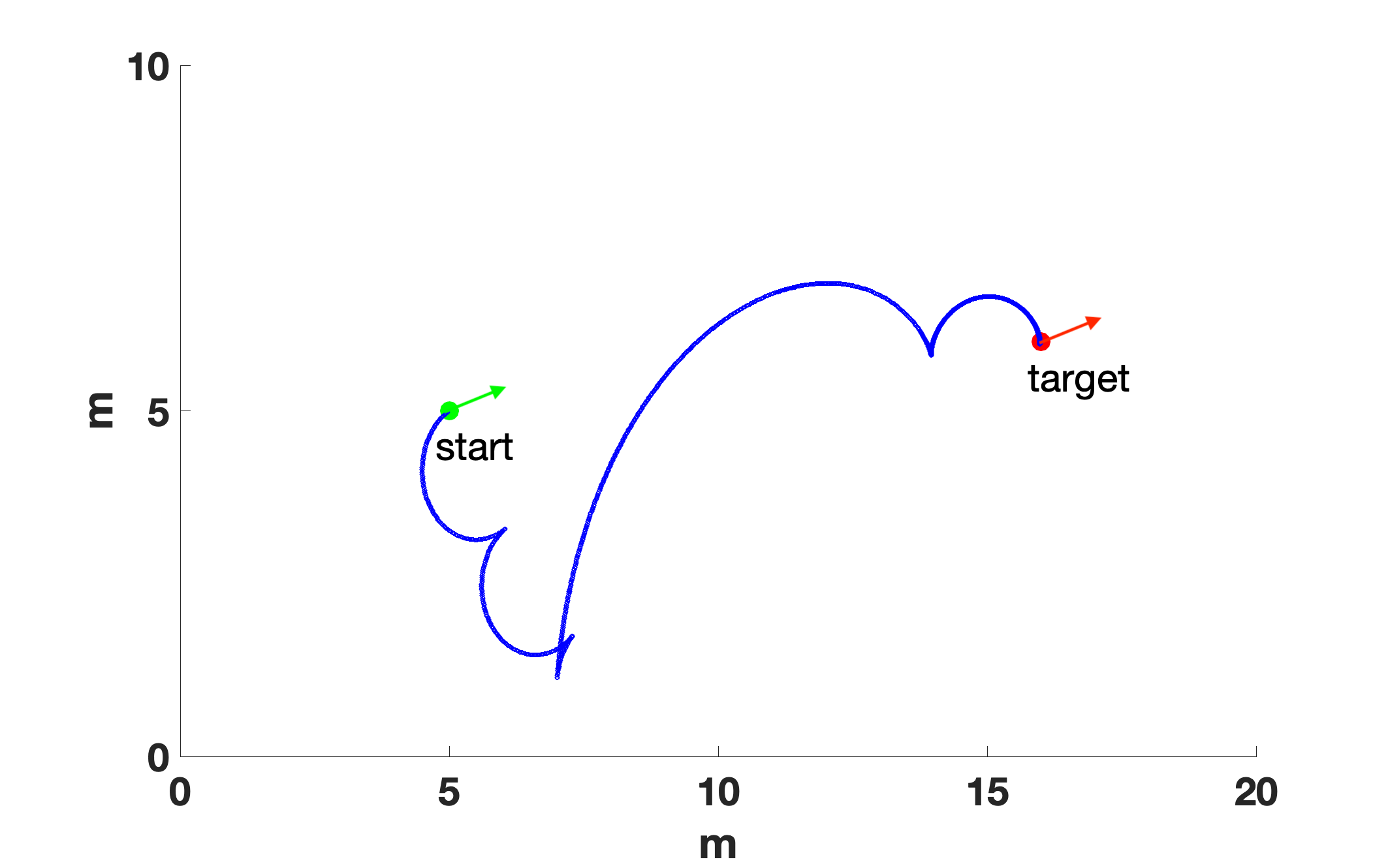}}
	\subfigure[]
	{\includegraphics[width=1.65in]{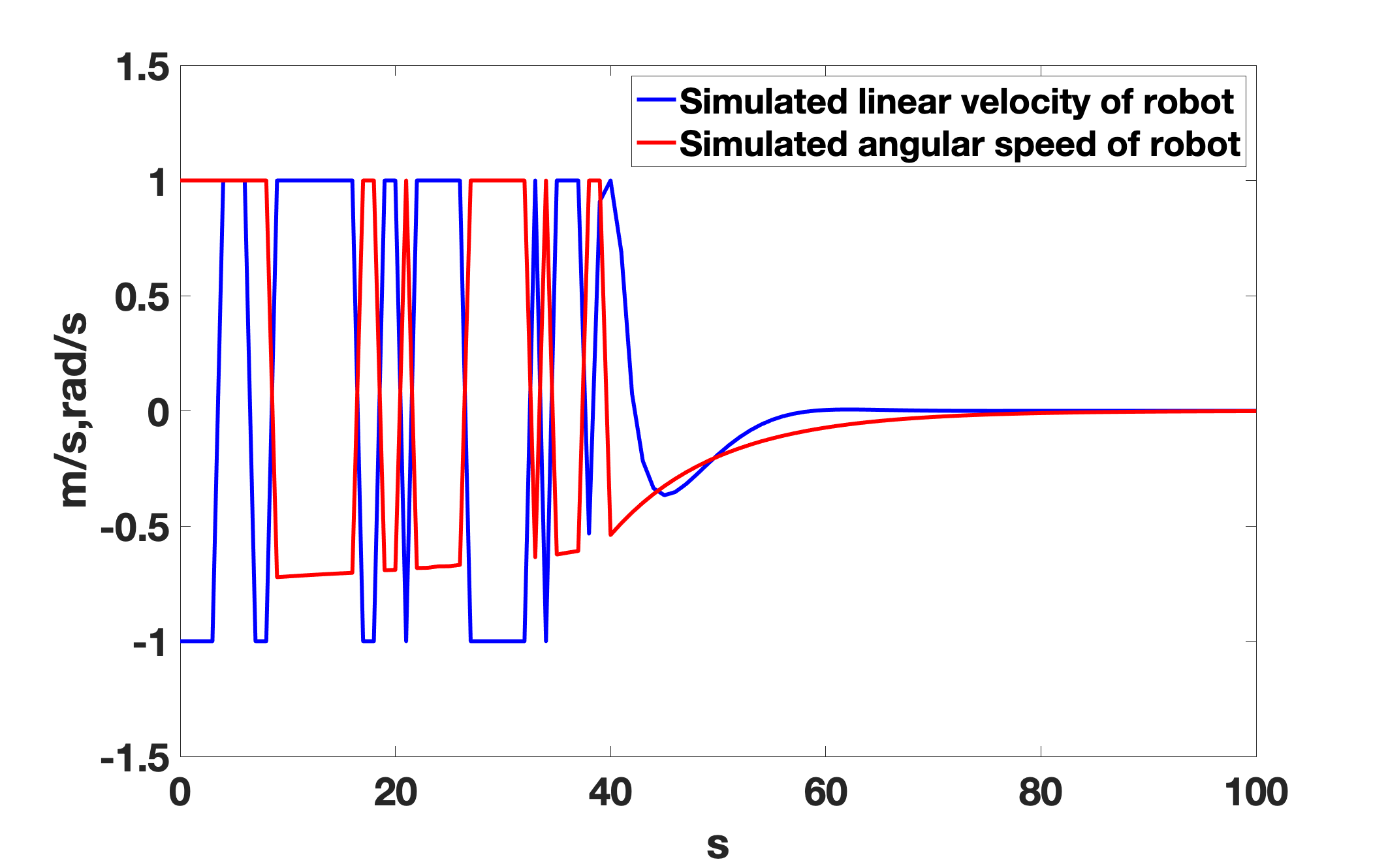}}
	\subfigure[]
	{\includegraphics[width=1.65in]{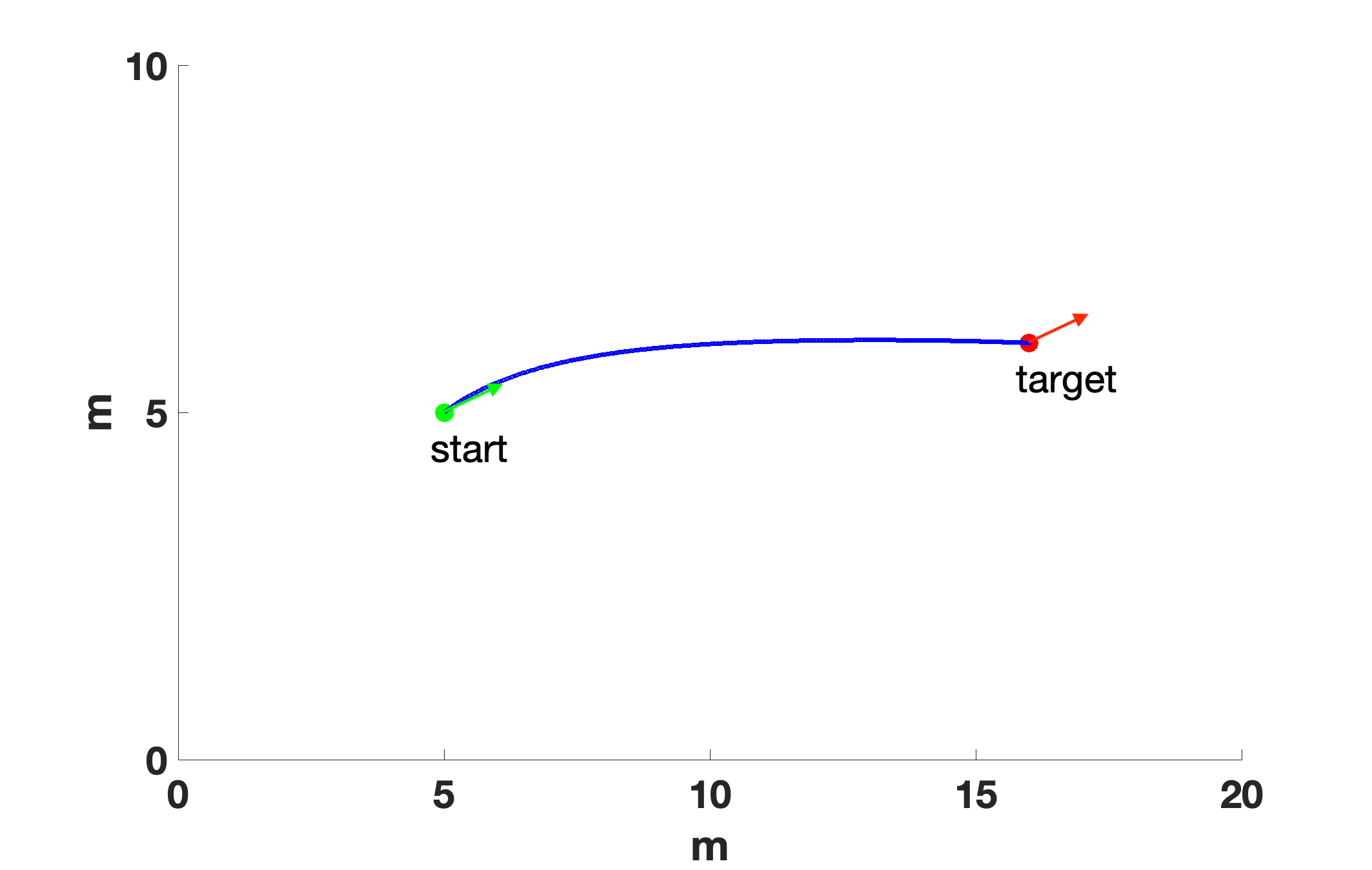}}
	\subfigure[]
	{\includegraphics[width=1.65in]{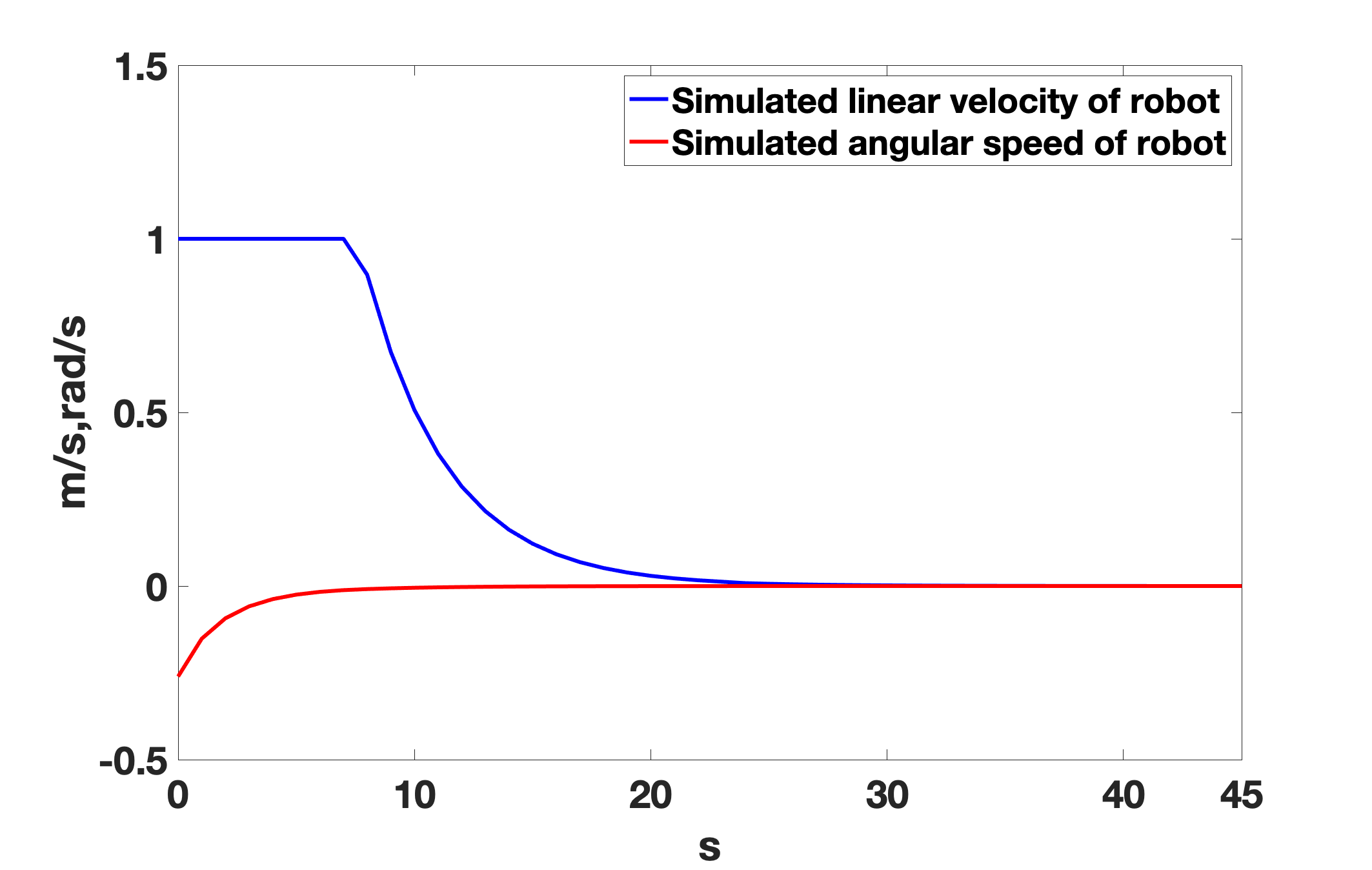}}
	\subfigure[]
	{\includegraphics[width=1.65in]{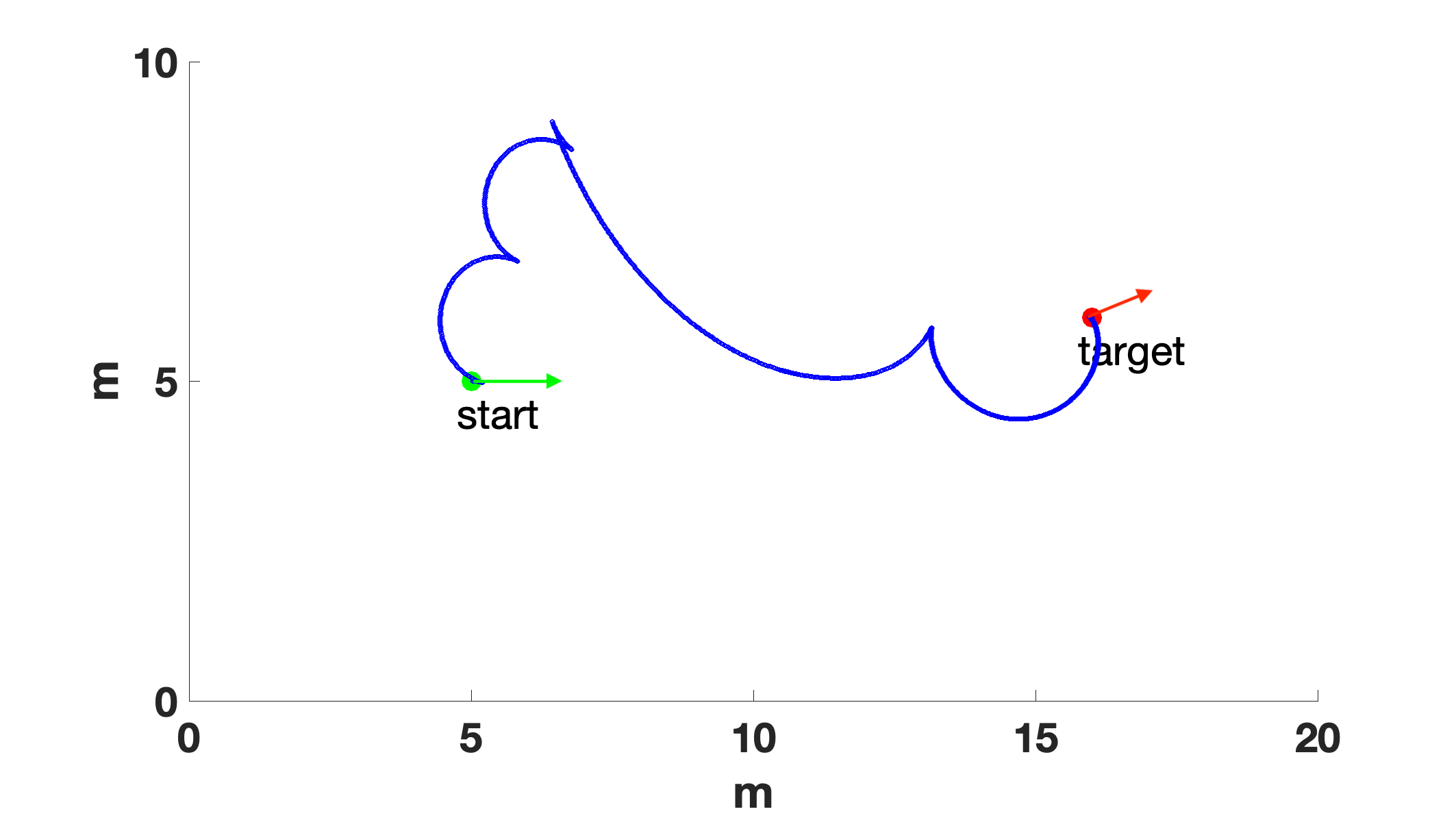}}
	\subfigure[]
	{\includegraphics[width=1.65in]{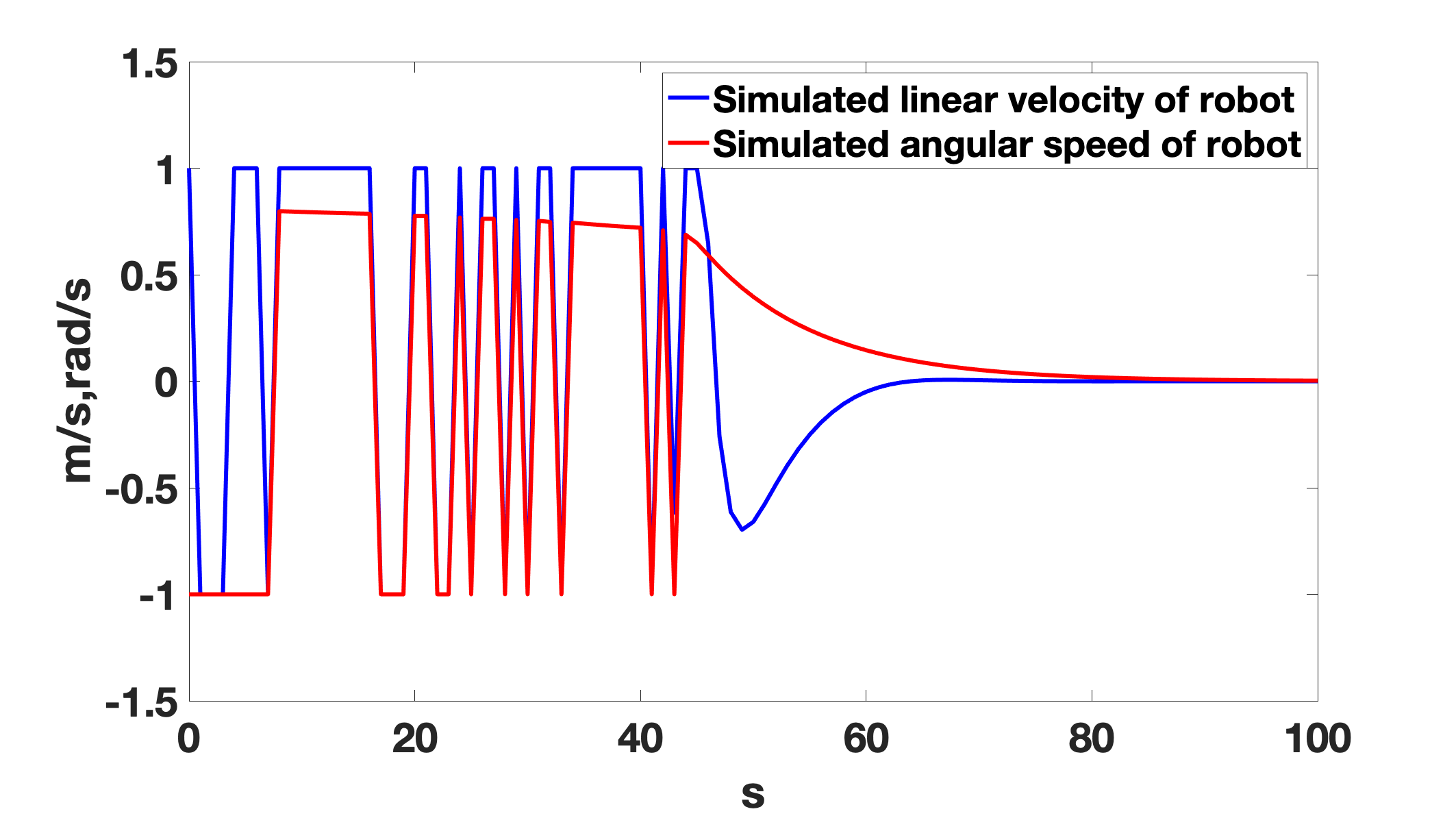}}
	\subfigure[]
	{\includegraphics[width=1.65in]{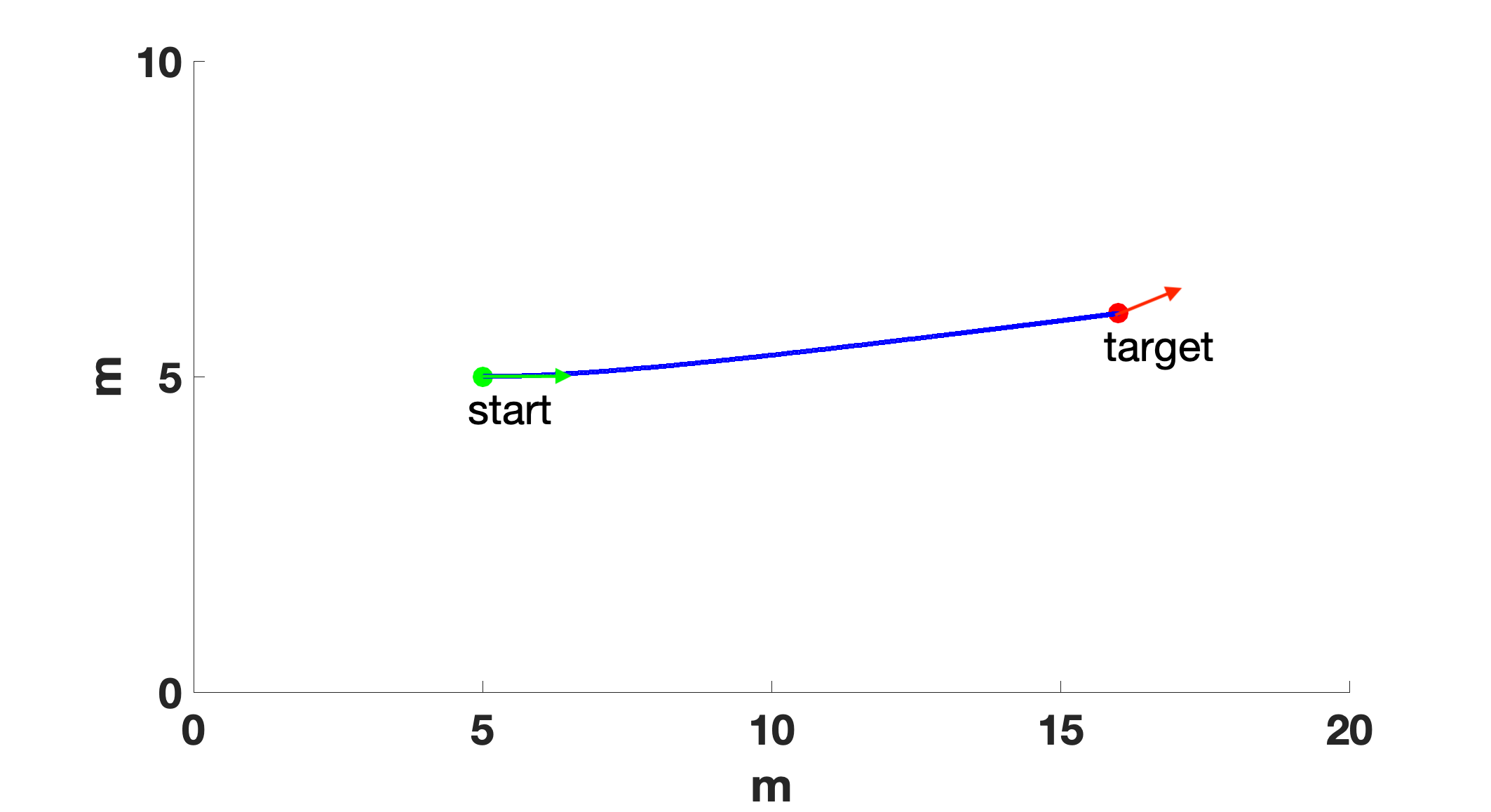}}
	\subfigure[]
	{\includegraphics[width=1.65in]{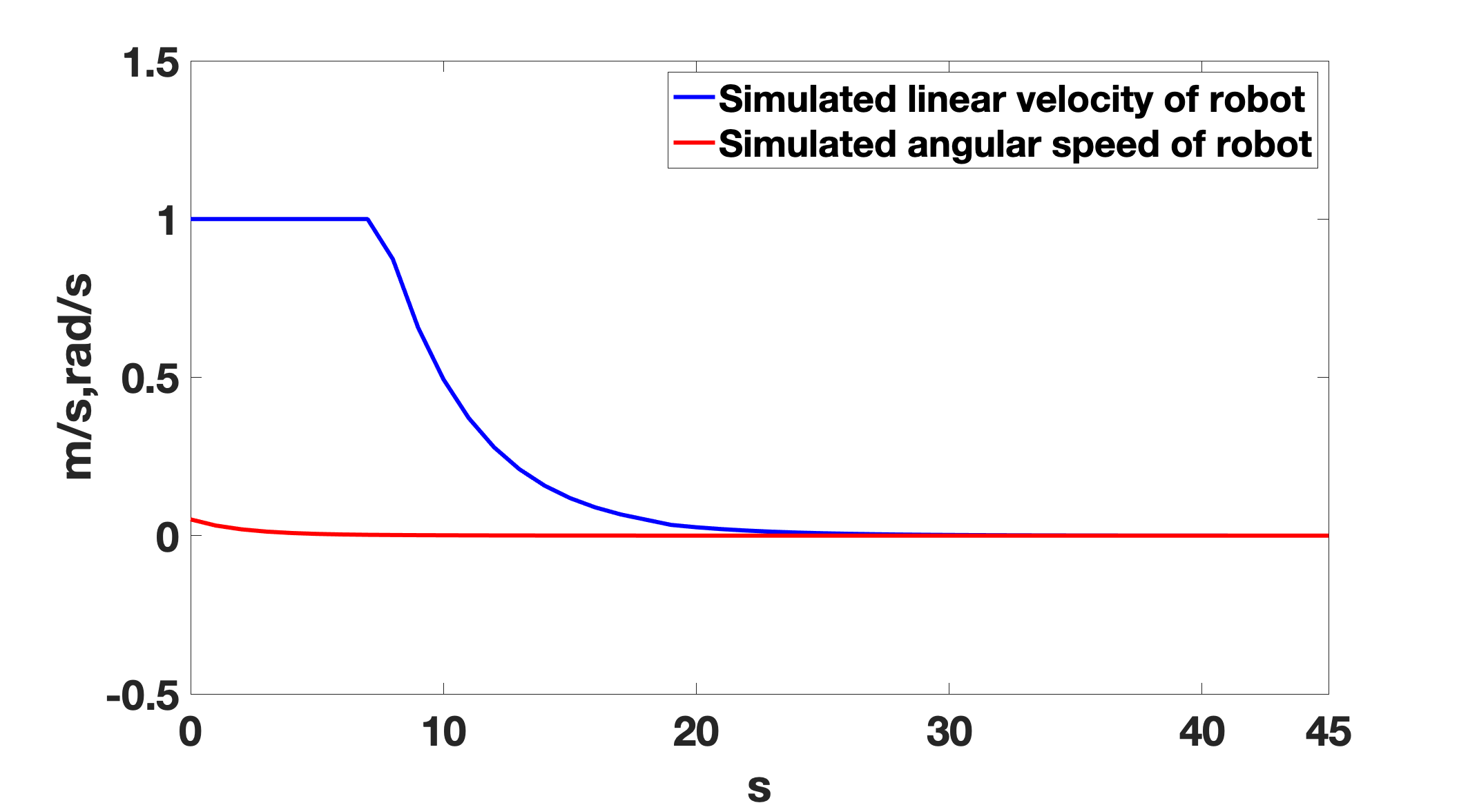}}
	
	\caption{Simulation trajectories (left) and control outputs (right) using the proposed controller and the controller in \cite{1marc2003} with kinematic constraints. (a)
		and (b): Trajectory and output by the controller in \cite{1marc2003} for Case 3; (c) and (d): Trajectory and output by the proposed controller for Case 3; (e) and (f): Trajectory and output by the controller in \cite{1marc2003} for Case 4; (g) and (h): Trajectory and output by the proposed controller for Case 4.}
	\label{figwithkc}
	\vspace{-0.1cm}
\end{figure*}

\begin{table}[!h]
	\caption{POSES SETTING FOR DIFFERENT CASES}
	\label{table}
	\begin{center}
		\tabcolsep 0.04in
		\renewcommand{\arraystretch}{1.5}
		\begin{tabular}{|c|c|c|c|c|}
			\hline
			Case & Case1 & Case2 & Case3 &  Case4 \\
			\hline
			Initial pose& $(0, 0, \pi/6)$  & $(0, 0, \pi/4)$  & $(5, 5, \pi/6)$  &  $(5, 5, 0)$  \\
			\hline
			Target pose& $(5, 5, 0)$  & $(5, 5, 0)$  & $(16, 6, \pi/6)$  &  $(16, 6, \pi/6)$  \\
			\hline
		\end{tabular}
	\end{center}
	\label{posecase}
\end{table}

For Case 1, it is demonstrated by  Fig. \ref{figwithoutkc} (a) that the singularity set is
indeed being escaped from by using the controller in \cite{1marc2003}.
When the proposed controller is employed, it can be observed from
by Fig. \ref{figwithoutkc} (c) that the robot is kept in the set Ss,c, resulting in much smoother and shorter trajectory. The linear velocity and angular velocity required by the proposed controller are much smaller than those required
by the existing controller as well. For Case 2,  the initial orientation is set to denoted by the green “arrow” , while the target pose remains unchanged in Fig. \ref{figwithoutkc} (g), the proposed controller is able to stabilize the robot at target pose with less control effort compared to the controller in \cite{1marc2003}.
In order to apply robot in some  practical situations, for Case 3 and Case 4, the linear and angular velocity of the robot are set to be bounded by 1 m/s and 1 rad/s, respectively. The
simulation results in Fig. \ref{figwithkc} are consistent with the previous
case. It can be observed that the robot is able to approach the target poses using both controllers. However, it is illustrated by Fig.  \ref{figwithkc} (c) and (g) that the proposed controller not only results in a shorter object servoing trajectory, but also doesn't cause  oscillating
behaviours as shown in Fig.  \ref{figwithkc} (b) and (f). In addition, it is worth noting that a parallel object servoing  task can be efficiently performed by the proposed controller in Case 3.
For Case 1 and Case 2, the curves of $|z_0|+|z_1|$ are shown in Fig. \ref{figcontroloutput}.  Normally, there are more overlapping area between a query image and the reference image if the value of $(|z_0|+|z_1|)$ is smaller. The value of $(|z_0|+|z_1 |)$ decreases sharply to a small value using the proposed controller  while it increases first before decreasing using the controller in \cite{1marc2003}. The simulation results indicate that the proposed controller  approaches the controllable ``singularity line", while the controller in \cite{1marc2003} tries to escape from the singularity set during the parking process. As a result, the proposed controller  is much more object servoing friendly than the controller in \cite{1marc2003}.

\section{Conclusion and Future Works}
\label{conclusion}
A novel object servoing scheme has been designed in this paper for differential-drive robots using visual based motion estimation and an object servoing friendly parking controller. The proposed scheme can be adopted to park a differential-drive robot at a predefined relative pose  to  a movable object. Due to the low on-line computational cost of the proposed scheme, it could be adopted for the last mile delivery of mobile robots to movable objects. It should be pointed out that the proposed scheme might not perform well in very cluttered environments.

\end{document}